\documentclass[conference]{IEEEtran}
\IEEEoverridecommandlockouts
\usepackage{cite}
\usepackage{amsmath,amssymb,amsfonts}
\usepackage{algorithmic}
\usepackage{graphicx}
\usepackage{textcomp}
\usepackage{xcolor}
\def\BibTeX{{\rm B\kern-.05em{\sc i\kern-.025em b}\kern-.08em
    T\kern-.1667em\lower.7ex\hbox{E}\kern-.125emX}}

\usepackage{times}
\usepackage{epsfig}
\usepackage{graphicx}
\usepackage{amsmath}
\usepackage{amssymb}
\usepackage{graphicx}
\usepackage{lscape}
\usepackage{comment}
\usepackage{caption}
\usepackage{subcaption}
\usepackage{float}
\usepackage{multirow}
\usepackage[utf8x]{inputenc}
\usepackage[breaklinks=true,bookmarks=false]{hyperref}
\begin{document}

\title{Now You See Me: Robust approach to Partial Occlusions}

 \author{ Karthick Gunasekaran \\
 University of Massachusetts\\
 Amherst, USA\\
 {\tt\small kgunasekaran@umass.edu}
 \and
 Nikita Jaiman\\
 University of Massachusetts\\
 Amherst, USA\\
 {\tt\small njaiman@umass.edu}

 }

\maketitle
\begin{abstract}
Occlusions of objects is one of the indispensable problems in Computer vision. While Convolutional Neural Networks (CNNs) provide various state of the art approaches for regular image classification, they however, prove to be not as effective for the classification of images with partial occlusions. Partial occlusion is scenario where an object is occluded partially by some other object/space. This problem when solved, holds tremendous potential to facilitate various scenarios. We in particular are interested in autonomous driving scenario and its implications in the same. Autonomous vehicle research is one of the hot topics of this decade, there are ample situations of partial occlusions of a driving sign or a person or other objects at different angles. Considering its prime importance in situations which can be further extended to video analytics of traffic data to handle crimes, anticipate income levels of various groups etc.,  this holds the potential  to be exploited in many ways. In this paper, we introduce our own synthetically created dataset by utilising Stanford Car Dataset and adding occlusions of various sizes and nature to it. On this created dataset,  we conducted a comprehensive analysis using various state of the art  CNN models such as VGG-19, ResNet 50/101, GoogleNet, DenseNet 121.  We further in depth study the effect of varying occlusion proportions and nature on the performance of these models by fine tuning and training these from scratch on dataset and how is it likely to perform when trained in different scenarios, i.e., performance when training with occluded images and unoccluded images, which model is more robust to partial occlusions and so on.
   
\end{abstract}

\section{Introduction}
Classification has proved to be a problem of prime significance in varied domains be it healthcare, autonomous driving or any other area and undeniably convolutional neural networks (CNNs) based architectures are the present  state of the art approach in this. When CNNs are reaching leaps and bounds in normal image classification \cite{krizhevsky2012imagenet}, these are relatively inefficient when classifying the objects with occlusions[occ]. Occ [Occlusions] have become quite challenging for the field of computer vision and hold a tremendous potential to uncover various benefits such as face detection, vehicle detection etc. 

Being motivated by it's scope, in the project, we are restricting to the problem of occ in the vehicular dataset.The dataset we consider is from Stanford called the "Cars Dataset"\cite{stanfordcar} having approx 16k varied  car images  with 196 categories of cars . The bigger picture that intrigued us is thinking about how accurately an autonomous self driving car of the future will be able to detect a human or a traffic sign or another vehicle with partial occs if they are using one of the state of the art of model.  When an object is on the road it is very highly likely to be occluded  by several objects we therefore, taking into consideration the impact it can have, try to focus on understanding and solving the problem of Partial Occs in vehicular datasets. 
We investigate how to build models resulting in better classification accuracies on the occluded dataset. For our purpose, an entire synthetic dataset is generated from the existing dataset with varied proportions and types of occ and conduct a detailed analysis by training different state of the art CNN Models which are either pre trained (on ImageNet \cite{deng2009imagenet}) or training the models from scratch for the classification of partially occ cars.

 The dataset is quite diverse and covers different car views from different angles. Classes are typically at the level of Make, Model, Year, e.g. 2012 Tesla Model S or 2012 BMW M3 coupe. 
The project is mainly a two step process: First, like discussed above, creating an artificially occluded dataset with variants.
Second, our goal is to evaluate various state of the art CNN architectures to observe the performance on occluded  data as compared to the unoccluded  and explore the ways to improve the classification accuracies by conducting various in detail experiments.

Through this paper we aim to investigate and answer several questions like, i) how occ sizes affects the performance of model architectures; ii) does deeper networks tend to perform better in the classification of partially occluded datasets; iii) robustness of model in the classification when trained on  dataset with different occ proportion and tested of different.
iv) would nature of artifact such as constant pixel artifact affect model performance as compared to random pixel artifact on image of same occ size and many more by conducting experiments of varied kinds extensively.


\section{Background/Related Work}
A good amount of work has been done on fine grained categorization of different objects such as birds \cite{farrell2011birdlets}, plants etc. already.  Fine grained categorization learns the sub details in one class as opposed to regular classification. Liu et al. \cite{liu2017monza} explains the fine grained classification techniques on the Stanford Car Dataset \cite{stanfordcar} identifying the make and models of cars from different angles. Owing to the fact that there is lack of huge dataset specific to cars, Yang et al. \cite{yang2015large} collects a large scale dataset called "CompCars" to perform fine grained categorization and verification of these cars on the dataset of both survellience nature and web nature containing approximately 200k images of 1.7 k car models suggesting interesting applications based on CNN models which can be made available on request.

 With these findings in the varied car dataset domain, there has been some work done in the domain of partial occ's as well. One challenging problem in the domain of vision is the problem of occs. Opitz et al. \cite{opitz2016grid} demonstrate in the domain of face detection their method outperforms the standard CNN models by introducing a new loss layers for CNN. How CNNs prove to be less effective in classification of occluded objects when they provide state of the art results for classifying an image without it, is talked about in Pepik et al. \cite{pepik2015holding} where they propose that one  key step to improve classification is by making the architectural changes enhancing the robustness to occs.
 
 Osherov et al. \cite{osherov2017increasing} discuss the relative insufficiency of convolutional neural networks in classification of partially occluded objects as compared to regular classification suggesting an alternate approach to enhance the robustness of CNNs to occs. They establish the finding that smaller filter size prove to be beneficial for the model's robustness to occs and propose an algorithm with new special regularisation to design a classifier that results in  robustness towards occs. 

\section{Approach}
\subsection{Custom Dataset Generation:}
\begin{figure}[H]
     \centering
     \begin{subfigure}[b]{0.13\textwidth}
           \centering
         \includegraphics[width=\textwidth]{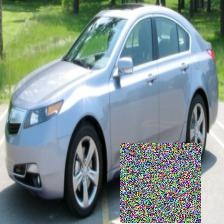}
         \caption{13\% }
         \label{fig:datasetapproachprop131}
     \end{subfigure}
     \centering
     \begin{subfigure}[b]{0.13\textwidth}
           \centering
         \includegraphics[width=\textwidth]{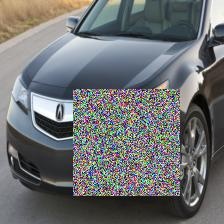}
         \caption{23\% }
         \label{fig:datasetapproachprop132}
     \end{subfigure}
     \centering
      \begin{subfigure}[b]{0.13\textwidth}
           \centering
         \includegraphics[width=\textwidth]{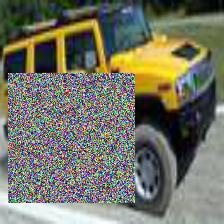}
         \caption{33\%}
         \label{fig:datasetapproachprop133}
     \end{subfigure}
       \caption{Different Occlusion Proportion}
        \label{fig:datasetapproachprop}
\end{figure}

One of the key challenges of the project was to create our own partially occluded dataset, since no such kind is available. Therefore, on the existing Stanford Car Dataset we artificially augmented partial occs of different proportions as demonstrated in figure \ref{fig:datasetapproachprop} to create the synthetic dataset. We also demonstrate the behaviour and performances of different model architectures with not just the occ size but also demonstrate what kind of difference will it make to have varied artifacts such as random pixel (type 1),constant pixel (type 2) or real images (type 3).

The dataset  contained lot of variations in images such as images with different angles,zoom levels, varied backgrounds etc. Data processing was carried to standardize the images such that occ size will remain uniform across images. In data processing step all the bounding boxes of images were  first extracted. The images which had bounding boxes sizes less than 224*224 were discarded so that upscaling doesn't affect the quality of the image.  Extracting the bounding boxes helps in standardizing the zoom level of objects in the images so that the occ ratio remains approximately the same across all the objects in image. Images were then down sampled to the standard size of 224 X 224. Post that artifacts were added to the images according to the size of occ. Images were then normalized and random horizontal flip was used as a data augmentation technique before being used by the model.

Data balancing was also the primary requirement,as the  dataset has inconsistent number of images in different classes. It was done to obtain even number of images across all the classes within threshold range. First, we divide the training and test data in a 70-30 \% split restricting the  number of images in each class within a particular range from the average (of total images). This reduces the likelihood of skew towards one particular class with the highest number of images. Next, we divide training data into 80-20 \% train-validation split ensuring each class has nearly equal representation in the validation dataset.


\subsection{Network Architectures:}
A variety of different architectures where studied and six were selected based on the how different they are from one another to perform our experiments on. However three variants of ResNet architecture were choosen to perform experiments on the depth of the network.
The network architectures chosen include VGG-19, GoogleNet \cite{googlenet}, DenseNet-121 \cite{densenet} and variants of ResNet \cite{resnet}; ResNet-50/101/151  for comparison of the model performance across all the combinations of occ proportions and artifact types. In total we trained eight state of the art models. The implementation was done in  PyTorch. We consider two types of optimizers; Adam  and Stochastic Gradient Descent (SGD) with momentum of 0.9. We observed that Adam is currently recommended as the default algorithm to use which certainly gives better performance for classification, therefore it was the first choice. However, we noticed that for our partially occluded dataset VGG-19 produced relatively lower accuracy with Adam hence, SGD + momentum was used as the optimizer with it.

We begin with two different techniques of weight initialisation i) using pre-trained weights on the ImageNet Dataset and ii) randomly initialising the weights to train it from scratch although we understand the repercussion of this approach owing to the small dataset size. Out of eight models, six of these we train using are pre-trained weights and other two from scratch.

When CNNs perform exceptionally on image classification, with respect to partial occ we intend to address following questions and findings such as 
\begin{itemize}
\item How performance of models is affected when trained on occluded dataset as compared to training on un-occluded dataset  when tested on partially occluded dataset?


\item Whether deeper networks are always able to learn finer image representations in the presence of occs as compared to simpler networks

\item Does the model performance change with varying occ size in training and testing or it remains invariant? 

 \item Will training the model from scratch, despite of having smaller dataset with data augmentation techniques perform better than pre-trained?  
 
\end{itemize}
First all the models where trained with different hyper parameters for different no of epochs.Then we experimented with increasing number of epochs realising that models from scratch would need higher epoch number to learn finer image representations. However, for pre-tained the fact that they are trained on large datasets there was no such necessity.

\section{Experiment}
\subsection{Dataset}

 \subsubsection{Cars Dataset}
Stanford Cars dataset \cite{stanfordcar} was used for this task. This dataset was choosen since one of the needs of any autonomous vehicles will be to identify objects which are partially occluded and cars will be one of main objects of interests for autonomous driving. As mentioned previously creating datasets was one of the main tasks of the project as there weren't any partially occluded datasets available before hand.

\subsubsection{Statistics} 
 For training and validation phase, we used \~10.3K images (details provided in table \ref{table:nonlin}) with 177 different classes of cars. Out of 10.3K images, approx  8k images where used for training and 2k images were used for validation. The images were evenly distributed across multiple classes with a mean of  59 images per classes and variance of 6 images across classes.  Also, we have divided the dataset in 70\%-30\% train-test split, respectively. For testing around 4k images where used.
  \begin{table}[H]
 \caption{Dataset Statistics} 
 \begin{tabular}{|l |c |  c|} 
 \hline
 Summary & Count\\ 
 \hline
Images In'Car Dataset' & 16k \\ 
\hline
Total classes & 196 \\
\hline
Total Occluded Images  & 14.8k \\
\hline
Classes with even distribution of images & 177 \\
\hline
Training images (\~70\%) & 10.5k \\
\hline
Train (\~80\%) -Val (\~20\%) Split & 8.3k - 2.1k \\
\hline
Test Images (\~30\%) & 4.3k\\
 \hline
 Occlusion proportions used & (13\%,23\%,33\%) \\
\hline
\end{tabular}
\label{table:nonlin}
\end{table}
 
 \subsubsection{Artifact types}
 Three different types of artifacts were generated to create varied datasets such that experiments can be performed to analyse the behaviour of models with respect to each artifact since in the real world, there are varying patterns of artifacts. The different artifact types are shown in figures \ref{fig:dataset13}, \ref{fig:dataset23} and \ref{fig:dataset33}. 


\begin{description}
\item[Type 1 Artifact:] 
 \hfill \\
This artifact was created by assigning random values between 0-255 for each pixel in all the images. Figure \ref{fig:dataset131} shows one such example of the random pixel artifact.

\item[Type 2 Artifact:] 
 \hfill \\
This artifact was result of using a constant value for all the pixels across images. Figure \ref{fig:dataset132} shows the constant pixel artifact.

\item[Type 3 Artifact:]
 \hfill \\
This was created by adding real life images on top of the images in the dataset. Figure \ref{fig:dataset133} shows the example of one such  artifact.
\end{description}

 \subsubsection{Occlusion Sizes}
 Datasets with varying Occ proportions were created to understand the effect of Occ sizes on the various model performances. The Occ sizes tested were 0\% (Un occluded), 13\%, 23\% and 33\%as shown below.
 
\begin{description}
\item[Un occluded dataset:]
 \hfill \\
 The unoccluded dataset was created to make a comparison with  occluded dataset  using various models as baseline is shown in Figure \ref{fig:dataset03}.
\begin{figure}[H]
     \centering
     \begin{subfigure}[b]{0.13\textwidth}
           \centering
         \includegraphics[width=\textwidth]{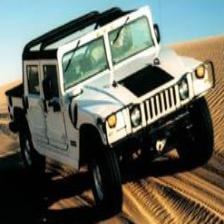}
         \caption{}
         \label{fig:dataset131}
     \end{subfigure}
     \centering
     \begin{subfigure}[b]{0.13\textwidth}
           \centering
         \includegraphics[width=\textwidth]{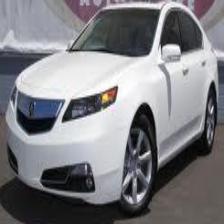}
         \caption{}
         \label{fig:dataset032}
     \end{subfigure}
     \centering
      \begin{subfigure}[b]{0.13\textwidth}
           \centering
         \includegraphics[width=\textwidth]{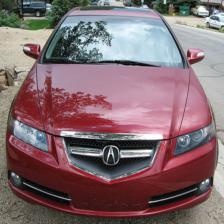}
         \caption{}
         \label{fig:dataset033}
     \end{subfigure}

       \caption{Unoccluded dataset}
        \label{fig:dataset03}
\end{figure}

 \item[13\% Occluded dataset:]
 \hfill \\
An artifact was added such that 13\% of the pixels in the original data was occluded. For this occ ratio three different types of artifacts where created as shown in    \ref{fig:dataset13}.

\begin{figure}[H]
     \centering
     \begin{subfigure}[b]{0.13\textwidth}
           \centering
         \includegraphics[width=\textwidth]{Images/Datasets/13/01.jpg}
         \caption{Type 1}
         \label{fig:dataset131}
     \end{subfigure}
     \centering
     \begin{subfigure}[b]{0.13\textwidth}
           \centering
         \includegraphics[width=\textwidth]{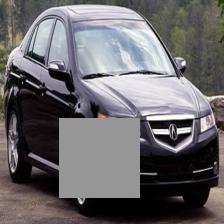}
         \caption{Type 2}
         \label{fig:dataset132}
     \end{subfigure}
     \centering
      \begin{subfigure}[b]{0.13\textwidth}
           \centering
         \includegraphics[width=\textwidth]{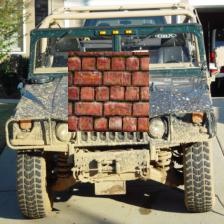}
         \caption{Type 3}
         \label{fig:dataset133}
     \end{subfigure}
       \caption{Different artifacts in dataset with 13\% occlusion on Images}
        \label{fig:dataset13}
\end{figure}

 \item[23\% Occluded dataset:]
 \hfill \\
Similar to previous one, artifact was created such that 23\% of the pixels in the original data was occluded. Three different types of artifacts were created for this occ ratio as shown in \ref{fig:dataset23}.

\begin{figure}[H]
     \centering
     \begin{subfigure}[b]{0.13\textwidth}
           \centering
         \includegraphics[width=\textwidth]{Images/Datasets/23/01.jpg}
         \caption{Type 1}
         \label{fig:dataset231}
     \end{subfigure}
     \centering
     \begin{subfigure}[b]{0.13\textwidth}
           \centering
         \includegraphics[width=\textwidth]{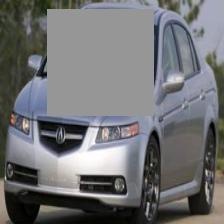}
         \caption{Type 2}
         \label{fig:dataset232}
     \end{subfigure}
     \centering
      \begin{subfigure}[b]{0.13\textwidth}
           \centering
         \includegraphics[width=\textwidth]{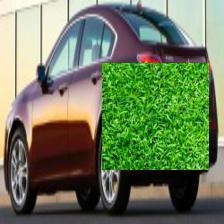}
         \caption{Type 3}
         \label{fig:dataset233}
     \end{subfigure}

       \caption{Different artifacts in datasets with 23\% occlusion}
        \label{fig:dataset23}
\end{figure}

 \item[33\% Occluded dataset:]
 \hfill \\
Artifact were created  in such a fashion that 33\% of pixels in the original image were occluded. Similar to previous data set,three different types of artifacts were created as shown in  Figure \ref{fig:dataset33}.
 
\begin{figure}[H]
     \centering
     \begin{subfigure}[b]{0.13\textwidth}
           \centering
         \includegraphics[width=\textwidth]{Images/Datasets/33/01.jpg}
         \caption{Type 1}
         \label{fig:dataset331}
     \end{subfigure}
     \centering
     \begin{subfigure}[b]{0.13\textwidth}
           \centering
         \includegraphics[width=\textwidth]{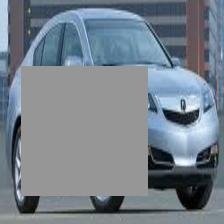}
         \caption{Type 2}
         \label{fig:dataset332}
     \end{subfigure}
     \centering
      \begin{subfigure}[b]{0.13\textwidth}
           \centering
         \includegraphics[width=\textwidth]{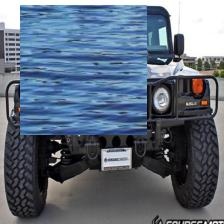}
         \caption{Type 3}
         \label{fig:dataset333}
     \end{subfigure}

       \caption{Different artifacts in datasets with 33\% occlusion}
        \label{fig:dataset33}
\end{figure}

\end{description}

\subsection{Evaluation Metric}
For the experiments conducted,top-1\%  and top-5\% accuracies were considered as an evalutation metric for  the model's performance. The top-1\% accuracy of the model is obtained by only considering the result with highest probability and comparing it with the ground truth label. The top-5\% accuracy  is taken by considering  5 highest probabilities in the predicted result and checking if one of them is the ground truth label.  

\subsection{Results}

\subsubsection{Effect of model trained with lower occluded data on higher occluded data:}

An interesting experiment was conducted to understand the behaviour of models when they come across higher occluded data but the training had only smaller sized occ. A model which was trained on 13\% Occ dataset was tested with 23\% and also 33\% Occ dataset. The baseline results for this case would be the model to be tested on same Occ size as trained i.e. baseline models trained on 13\% and 23\% will be tested on 13\% and 23\% respectively to be  compared. The results can be seen in table \ref{tab:exp5_tab}.

\begin{table}[H] 
\centering

 \caption{\label{tab:exp5_tab}Effect of training on Lower Occ Data and Testing on higher Occ Data} 
 
 \begin{tabular}{|l | c| c| c| c|} 
 \hline
  \multirow{2}{*}{Trained Occ} & \multirow{2}{*}{Tested Occ} & \multicolumn{2}{c|}{Model}   \\ 
 & & Top 1\% & Top 5\%  \\
 \hline
13\% & 23\% & 80.14  & 94.47 \\ 
\hline
13\% & 33\% & 70.62 &	91.03\\
\hline
23\% & 23\%  & 81.87 & 95.81 \\
\hline
 33\% & 33\% &  72.85 & 92.42  \\

\hline
0\%(Un-occ) & 23\% & 62.87 & 86.50\\

\hline
0\%(Un-occ) & 33\% & 45.68 & 74.17\\
\hline
\end{tabular}
\end{table}

\begin{figure}[H]
       \includegraphics[width=0.5\textwidth]{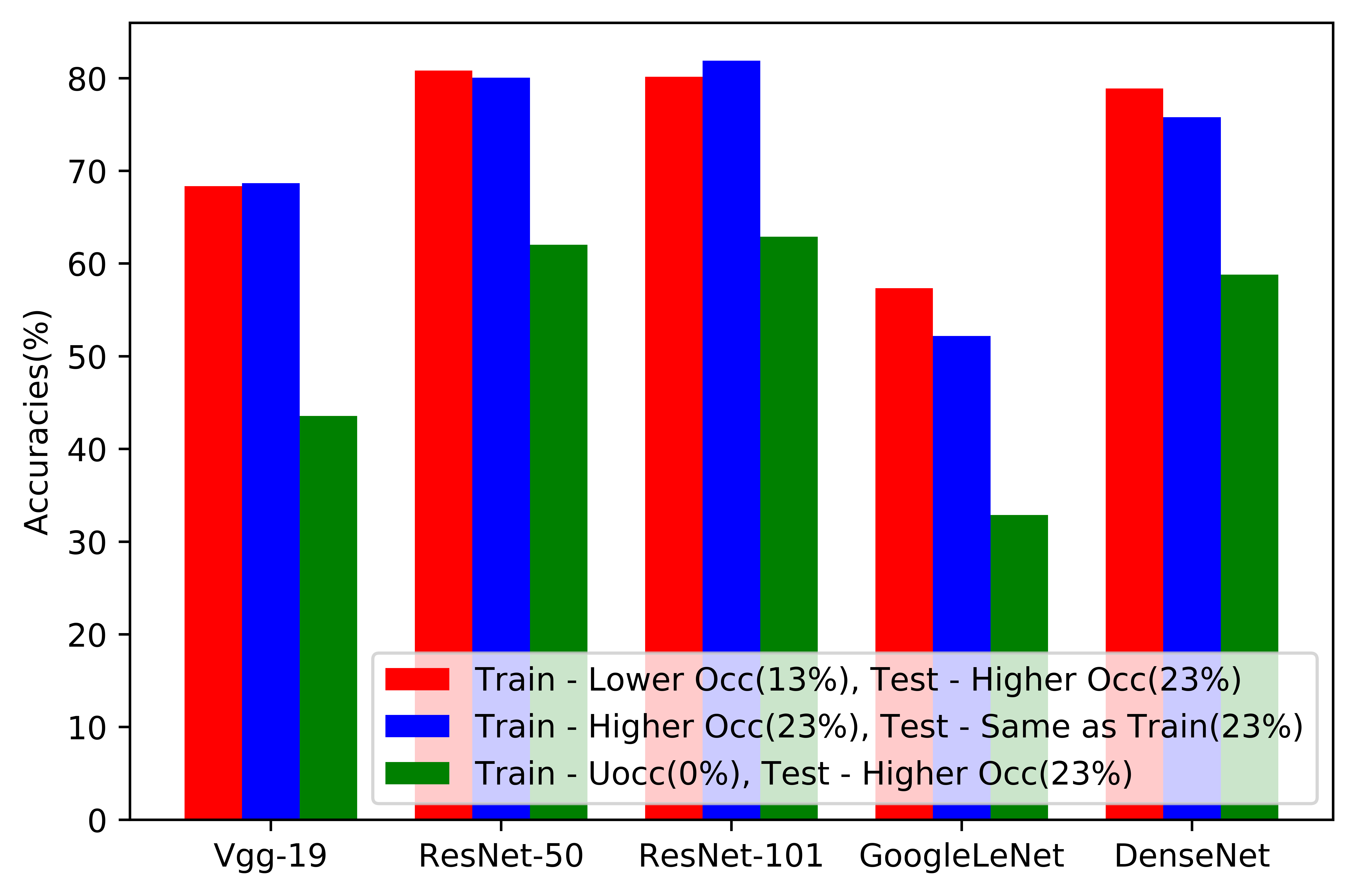}
        \caption{Analysis on Effect of model trained with lower occluded data on higher occluded data}
        \label{fig:lowerocctrainopp}
\end{figure}

From Table \ref{tab:exp5_tab} above, we can see that the models trained on lower Occ data gave similar performance when tested on the higher Occ data as compared to the models trained on higher Occ data when tested with same Occ size. Also, we note that the models trained on unoccluded dataset does not perform well. The Figure \ref{fig:lowerocctrainopp} clearly shows that across all the models same behaviour can be seen. 

Therefore, we understand that the models are unable to generalize to artifacts if they have not seen these before during the training time, however, are able to adapt and generalize well for larger artifacts if they have seen smaller artifacts previously.

\subsubsection{Effect of model trained with higher occluded data on unoccluded or lower occluded data :}
In this experiment the performance of models on lower Occ or un-occluded data when they were trained on higher Occ data was analysed. In the real world scenario, there will be cases where the training set has an object and testing set doesn't contain any occ. This test was carried out to understand if the model which learnt from lesser number of features because of occ would be able to perform similar to the model which was trained with all the features available. The baseline in this case is the model which was trained on un-occluded dataset and tested on un-occluded dataset since they have  all the features available to them during training and were able to classify pretty much accurately. 
   
\begin{table}[H] 
\centering

 \caption{\label{tab:exp7_tab}Effect of training on Higher Occ Datasets and Testing on Lower Occ Datasets} 
 
 \begin{tabular}{|l | c| c| c| c|} 
 \hline
   \multirow{2}{*}{Trained Occ} & \multirow{2}{*}{Tested Occ} & \multicolumn{2}{c|}{Model}   \\ 
 
 & & Top 1\% & Top 5\%  \\
 \hline

13\% & 0\% & 89.59  & 98.29 \\ 

\hline
23\% & 0\% & 88.77 &	97.95\\

\hline
0\% & 0\%  & 89.43 & 98.38 \\

\hline
 13\% & 13\% &  85.63 & 97.33  \\
\hline
23\% & 23\% & 81.87 & 95.81\\

\hline
\end{tabular}
\end{table}
The Table \ref{tab:exp7_tab}  shows the top-1 and top-5 accuracy of model when tested on lower occ datasets.
\begin{figure}[H]
       \includegraphics[width=0.5\textwidth]{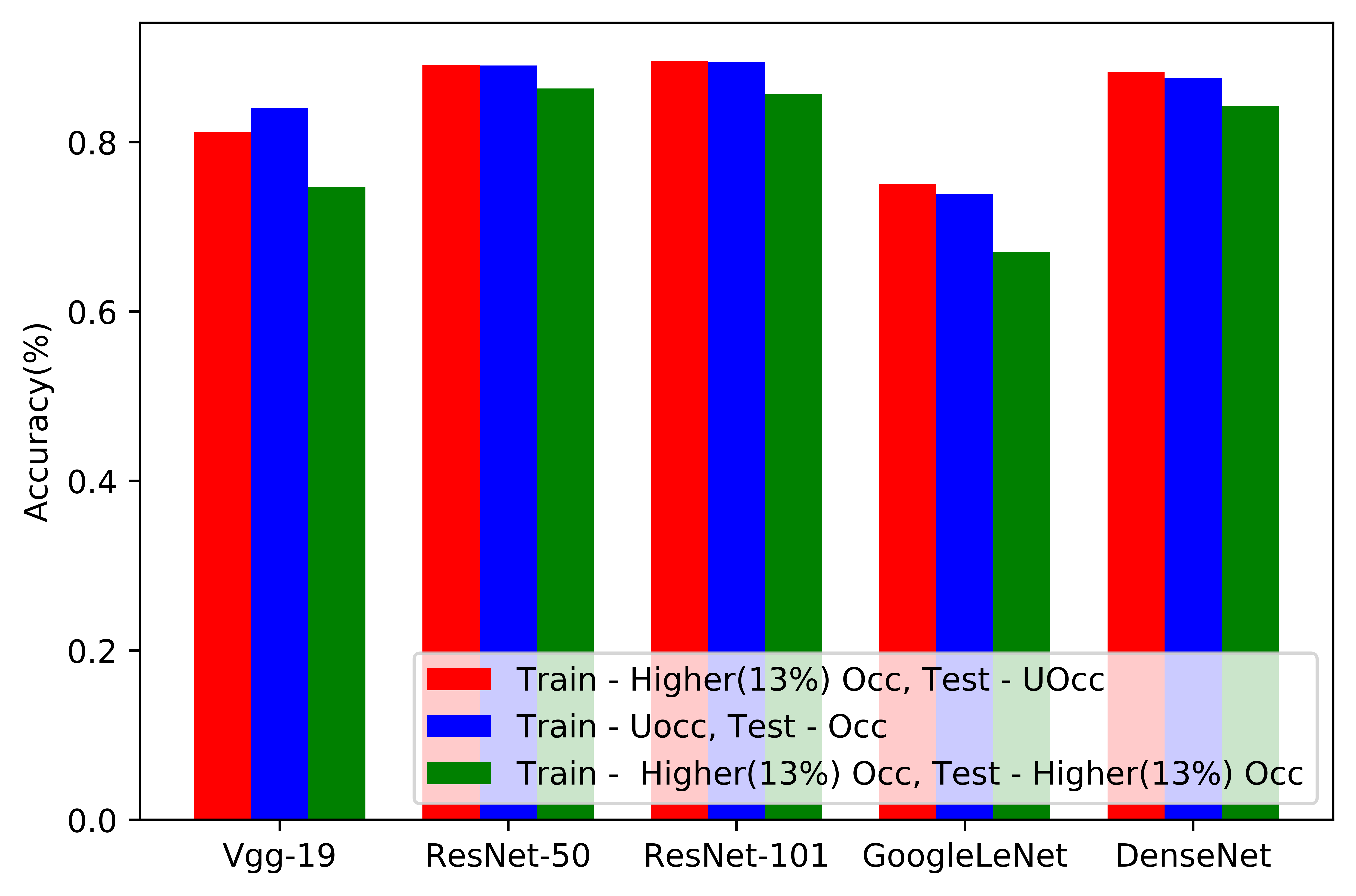}
        \caption{Analysis on Effect of model trained with higher occluded data on unoccluded or lower occluded data }
        \label{fig:higherocctrainopp}
\end{figure}
Figure \ref{fig:higherocctrainopp} shows that the model performance is almost similar to the baseline. Therefore, it can be inferred that having occluded objects in the training set will be an added advantage since it will be able to classify both occluded and un-occluded objects in test and better helps  in overall generalization of the model.

\subsubsection{Effect of different models on different Occlusion sizes}

Understanding the model performance with respect to occ sizes was one of the important aspect of this work. Therefore, models were tested for datasets with different occ percentages of Type 1 Artifact. All the models listed were pretrained on Imagenet and then again trained on Stanford Cars un-occluded dataset \cite{stanfordcar}. Table \ref{tab:exp1_tab} shows the top-1\% and top-5\% accuracies of the models when tested with different sizes of artifacts. 

\begin{table*}[ht] 
\centering

 \caption{\label{tab:exp1_tab}Model performance across different Occlusions Sizes} 
 
 \begin{tabular}{|l | c| c| c| c|c| c| c| c| c|} 
 \hline
\multirow{2}{*}{Model} & \multicolumn{2}{c|}{Unacc}  & \multicolumn{2}{c|}{13\%} & \multicolumn{2}{c|}{23\%} & \multicolumn{2}{c|}{33\%} \\ 
 
  & Top 1\% & Top 5\% & Top 1\% & Top 5\% & Top 1\% & Top 5\% & Top 1\% & Top 5\%\\
 \hline

Vgg-19 & 84.01 & 96.29 & 60.73 & 83.60 & 43.54 & 69.89 & 28.14 & 53.79 \\ 

\hline
ResNet-50 &  89.02 & 98.15 & 75.50 & 93.62 & 62.01 & 86.59 & 44.14 & 73.13  \\

\hline
ResNet-101   & 89.43 & 98.38 & 77.75 & 94.10 & 62.88 & 86.49 & 45.68 & 74.17  \\

\hline
 GoogleLeNet &  73.88 & 90.89 & 49.42 & 74.31 & 32.86 & 58.96 & 20.72 & 42.06 \\

\hline
DenseNet-121 & 87.57 & 97.56 & 74.50 & 92.46 & 58.80 & 84.22 & 40.86 & 67.59\\
\hline

\end{tabular}
\end{table*}

\begin{figure}[ht]
       \includegraphics[width=0.5\textwidth]{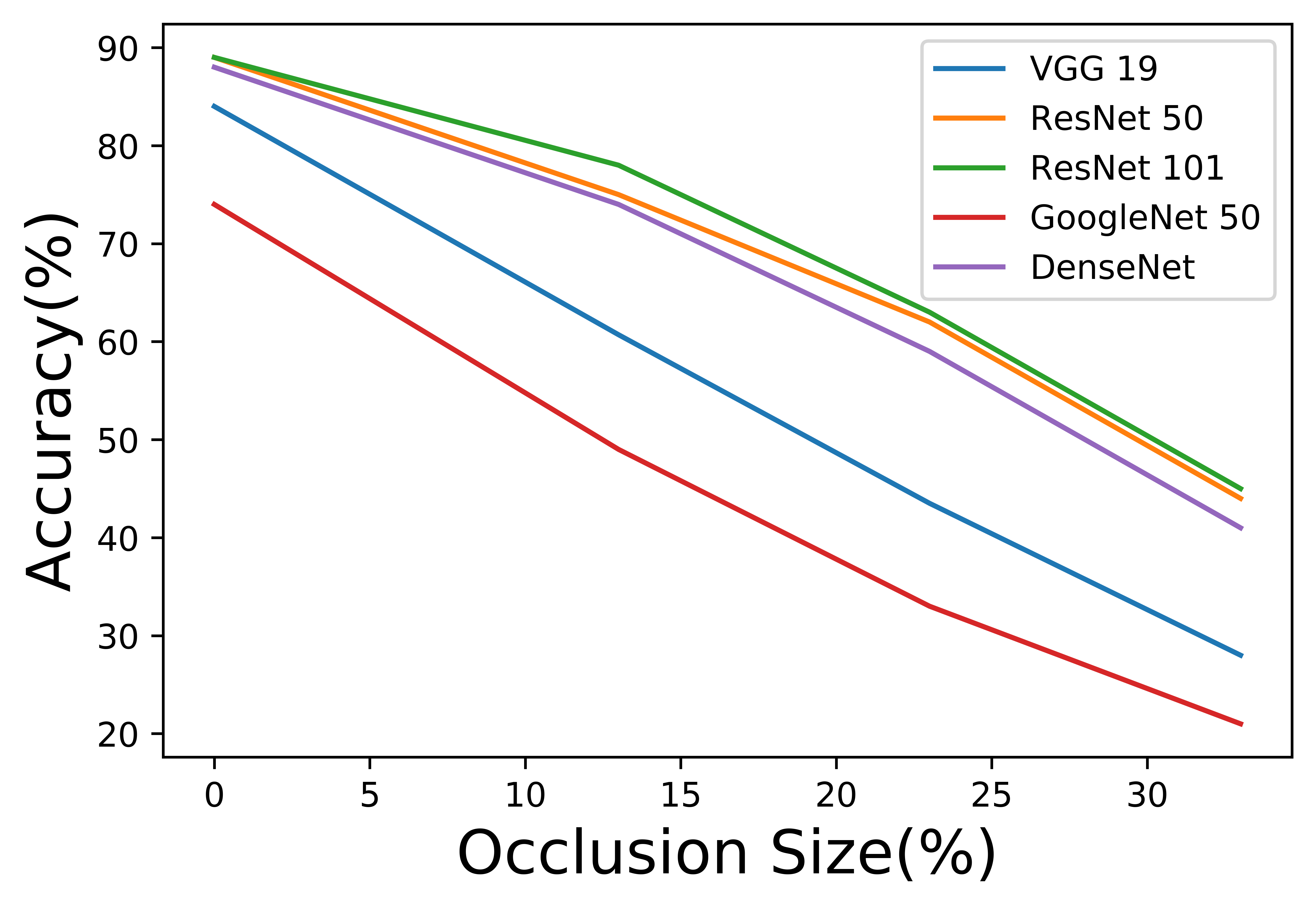}
        \caption{Accuracy[Top-1\%] of Models Vs Occlusion sizes}
        \label{fig:AccVsOccratio}
\end{figure}

From plot \ref{fig:AccVsOccratio} it can be seen that the accuracy of all the models decreases in almost a linear fashion as the occ sizes increase. Resnet-101,ResNet-50 and DenseNet tend to perform better overall for all types of data.  The models are unable to classify accurately as the occ sizes in the datasets increase. 

\begin{figure}[ht]
       \includegraphics[width=0.5\textwidth]{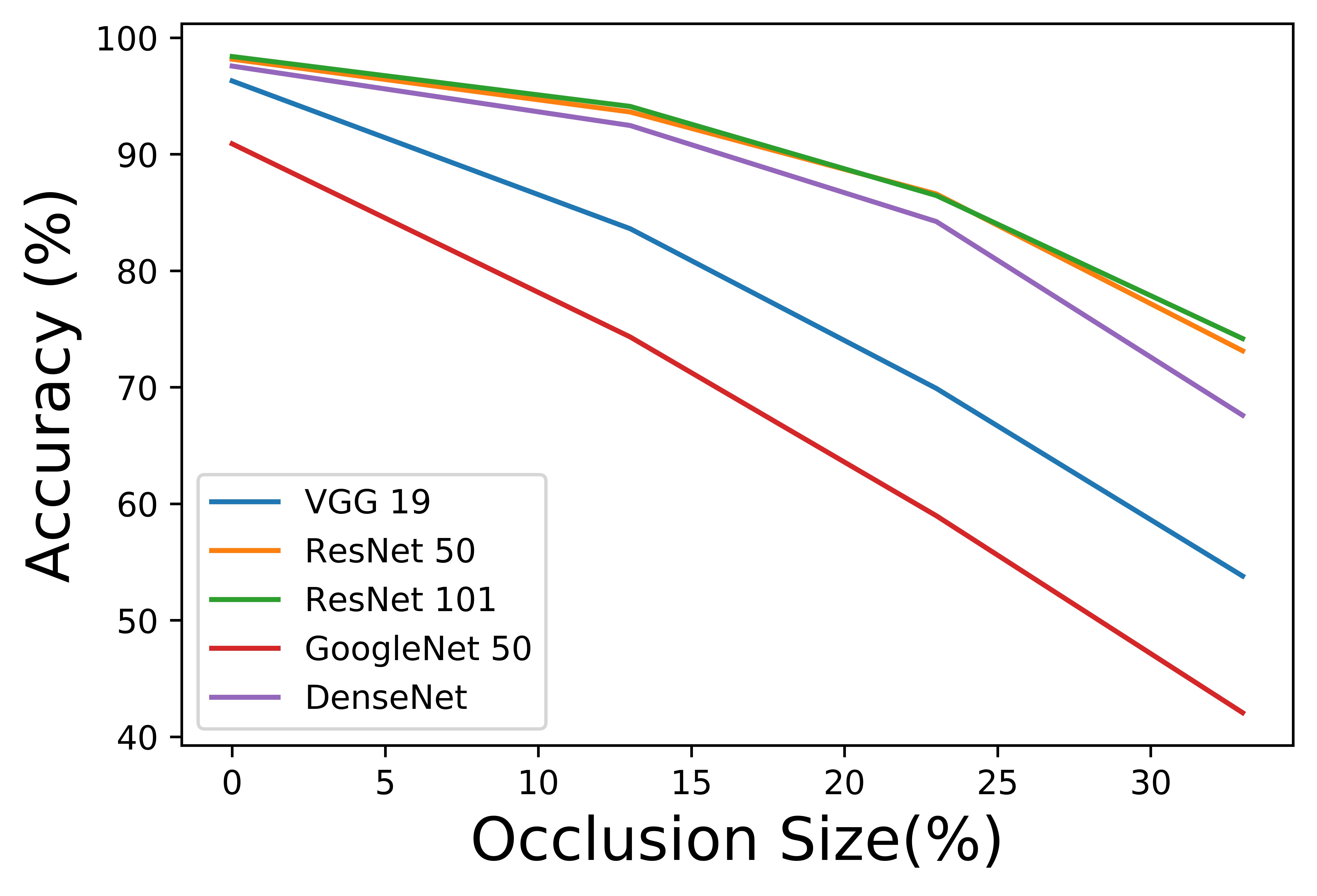}
        \caption{ Accuracy[Top-5\%] of Models Vs Occlusion sizes}
        \label{fig:AccVsOccratioTop5}
\end{figure}

 Further probing with top-5\% accuracy as shown in the Figure \ref{fig:AccVsOccratioTop5} it can be seen that Residual Network architecture in general is better compared with DenseNet, since a sudden decrease of top-5\% accuracy is seen in DenseNet which is not observed in ResNet. No specific effect was observed in models with respect to Occ, the performance of all the models degrade as the Occ size increases.      

\subsubsection{Effect of Training on occluded dataset and Training  on un-occluded dataset and testing Occluded data}
\begin{figure}[ht]
       \includegraphics[width=0.5\textwidth]{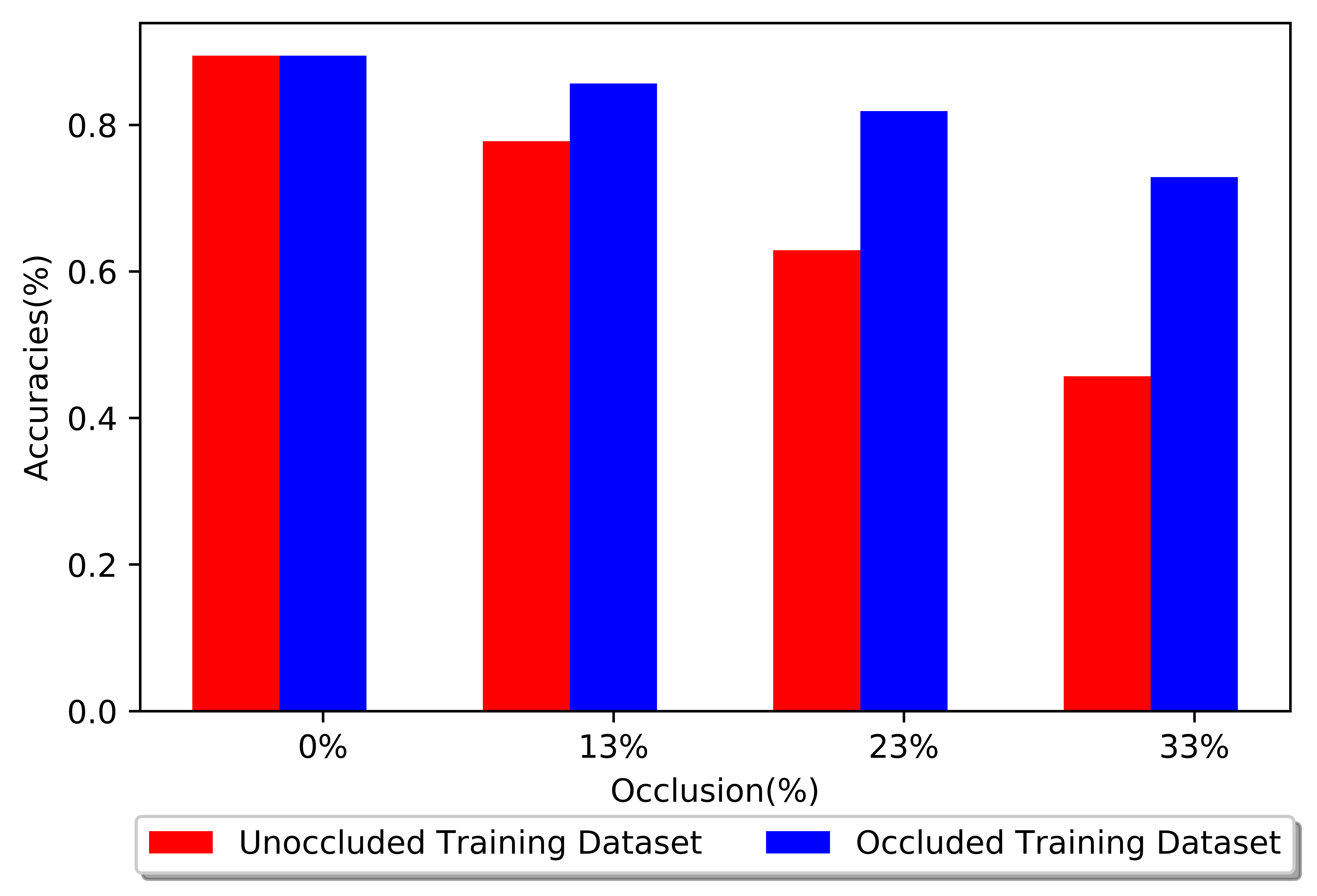}
        \caption{Effect of Training on Occlusion Dataset and Training  on un-occluded dataset for different Occlusions}
        \label{fig:effect_occluded_train}
\end{figure}

 A comparison was made to understand the effect of training the model with occluded training images. In this, a pretrained model was trained on occluded dataset and separately on un-occluded datasets. Now both the models were tested on the occluded dataset to determine how much better the model trained on occluded will perform. Results can be seen in the table \ref{tab:exp3_tab}. From the bar plot \ref{fig:effect_occluded_train} it can be noticed that as the occ size of the objects increases the performance of the model with Occ training performed much better than the one trained on un-occluded data. The difference between the performance of the model trained on occluded dataset and on un-occluded dataset was initially very low at 13\% occ size. In 23\% occ size, the difference between them increased further however at 33\% it increased to a very high value of 27\%  difference. The trend can be seen that model tend to perform better with training on occluded data. Similar kind of relationship can be seen with other models as well.  
 
\begin{table}[H] 
\centering

 \caption{\label{tab:exp3_tab}Model performance across different Occlusions Sizes} 
 
 \begin{tabular}{|l | c| c| c| c|c| c| c| c| c|} 
 \hline
  \multirow{2}{*}{Occ Size} & \multicolumn{2}{c|}{Unocc}  & \multicolumn{2}{c|}{Occ}  \\ 
  & Top 1\% & Top 5\% & Top 1\% & Top 5\% \\
 \hline
0\% & 89.43 & 98.38 & - & - \\ 
\hline
13\% & 77.75 &	94.10 &	85.63 &	97.34 \\
\hline
23\%   & 62.88 & 86.49 & 81.87 &	95.81 \\

\hline
 33\% &  45.68 & 74.17 & 72.85 & 92.42 \\
\hline
\end{tabular}
\end{table}

\subsubsection{Effect of Network depth on Occlusion}
\begin{figure}[ht]
       \includegraphics[width=0.5\textwidth]{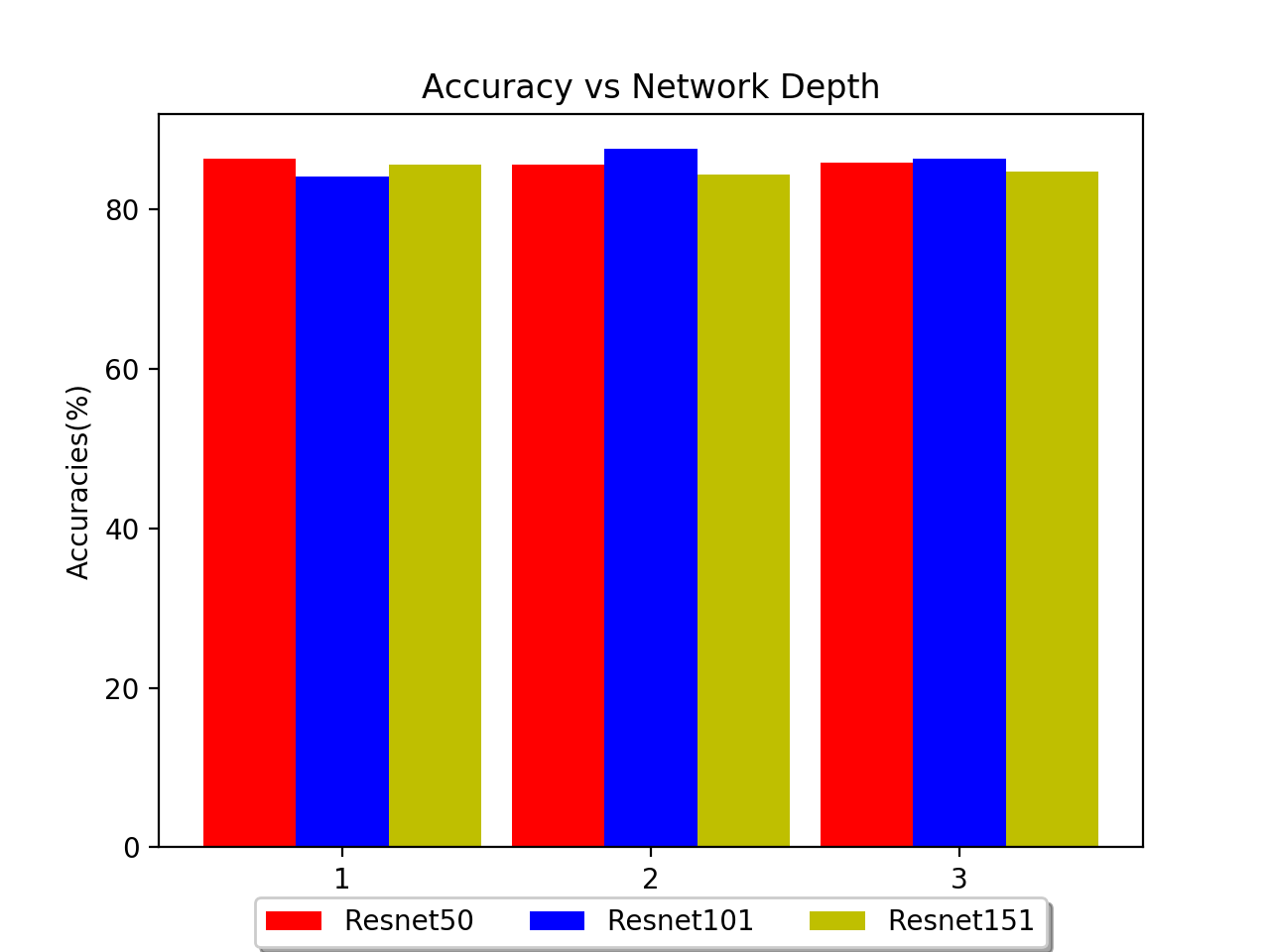}
        \caption{Accuracies(Top 1\%) with variation in network depth for artifact types 1,2 and 3}
        \label{fig:depth}
\end{figure}

This experiment was aimed at understanding the hypothesis how the network depth impacts the classification with partial occs. We consider different number of layers (50,101,151) of the same network architecture; Residual Network (ResNet) and investigate if increasing the number of layers in the network actually helps in understanding the finer image representations better with respect to occ than the one with lesser number of layers.
For the car dataset with 13\% occ(details mentioned in table \ref{table:table_depth}), we create a bar plot of for artifacts type 1,2 and 3 and plot the corresponding accuracies obtained by training the  ResNet-50, ResNet-101 and  ResNet-151 models
as shown in figure \ref{fig:depth}. The ResNet  architecture was chosen because it performed better comparatively than others.

As seen in the figure, it's quite clear that in case of partial occs increasing network depth doesn't play significant role in increasing or decreasing the accuracies. We conclude in case of partial occs, depth of the network doesn't seem to have impact on model performances in classification.

 \begin{table}[ht]
 \setlength{\tabcolsep}{2pt}
 \caption{Effect of network depth on model performance for different artifacts  with 13 \% occ} 
 \begin{tabular}{|l |c|c|c|c|c|c|c|} 
 \hline
 \multirow{2}{*}{Type} & \multicolumn{2}{c|}{ResNet-50} &\multicolumn{2}{c|}{ResNet-101}  & \multicolumn{2}{c|}{ResNet-151 }  \\ 
 & Top 1\% & Top 5\% & Top 1\% & Top 5\% & Top 1\% & Top 5\% \\
  \hline
1 & 86.31 &97.11 &85.63 &97.34 & 85.89 & 97.49 \\
\hline
2 & 84.13 &96.90& 87.54 &97.79& 86.35 &96.63\\
\hline
3 & 85.56 &97.06& 84.35 &96.95&84.69&96.45\\
 \hline
\end{tabular}
\label{table:table_depth}
\end{table}

\subsubsection{Comparison of un-pretrained model and pre-trained model performance on Occlusion}

As discussed above, there are two ways to train a model; one is to pretrain the model on Imagenet dataset and use the pre-trained weights to retrain on Stanford cars dataset \cite{stanfordcar} and  second take randomly initialized weights and train the model from scratch on Stanford cars dataset. For the car dataset with 13\% occs and artifact type 1 we considered two models to be trained by both using the pre-trained weights and from scratch.  Table \ref{table:ptnpt} gives the details of top 1\% and 5\% accuracies corresponding to these values. One of the primary reason that lead to poor performance of the models trained from scratch is due to the limited size of the dataset and large number of classes (177) even though we used few data augmentation techniques they proved to be insufficient here. So, the effect of Occ could not be determined in this case.  

\begin{table}[H]
 \caption{Pretrained (PT) vs Not pre-trained(NPT) models for 13\% occlusion and artifact type 1} 
 \begin{tabular}{|l |c |c|c |c|} 
  \hline
\multirow{2}{*}{Model} & \multicolumn{2}{c|}{PT} & \multicolumn{2}{c|}{N-PT}\\
  & Top 1\% & Top 5\% & Top 1\% & Top 5\% \\ 
 \hline
VGG-19 &74.68 & 92.55 & 08.22 & 26.44 \\ 
\hline
ResNet-50 &86.31 & 97.10 & 20.61 & 45.82 \\
\hline
\end{tabular}
\label{table:ptnpt}
\end{table}

\subsubsection{Effect of different artifact types}
\begin{figure}[ht]
       \includegraphics[width=0.5\textwidth]{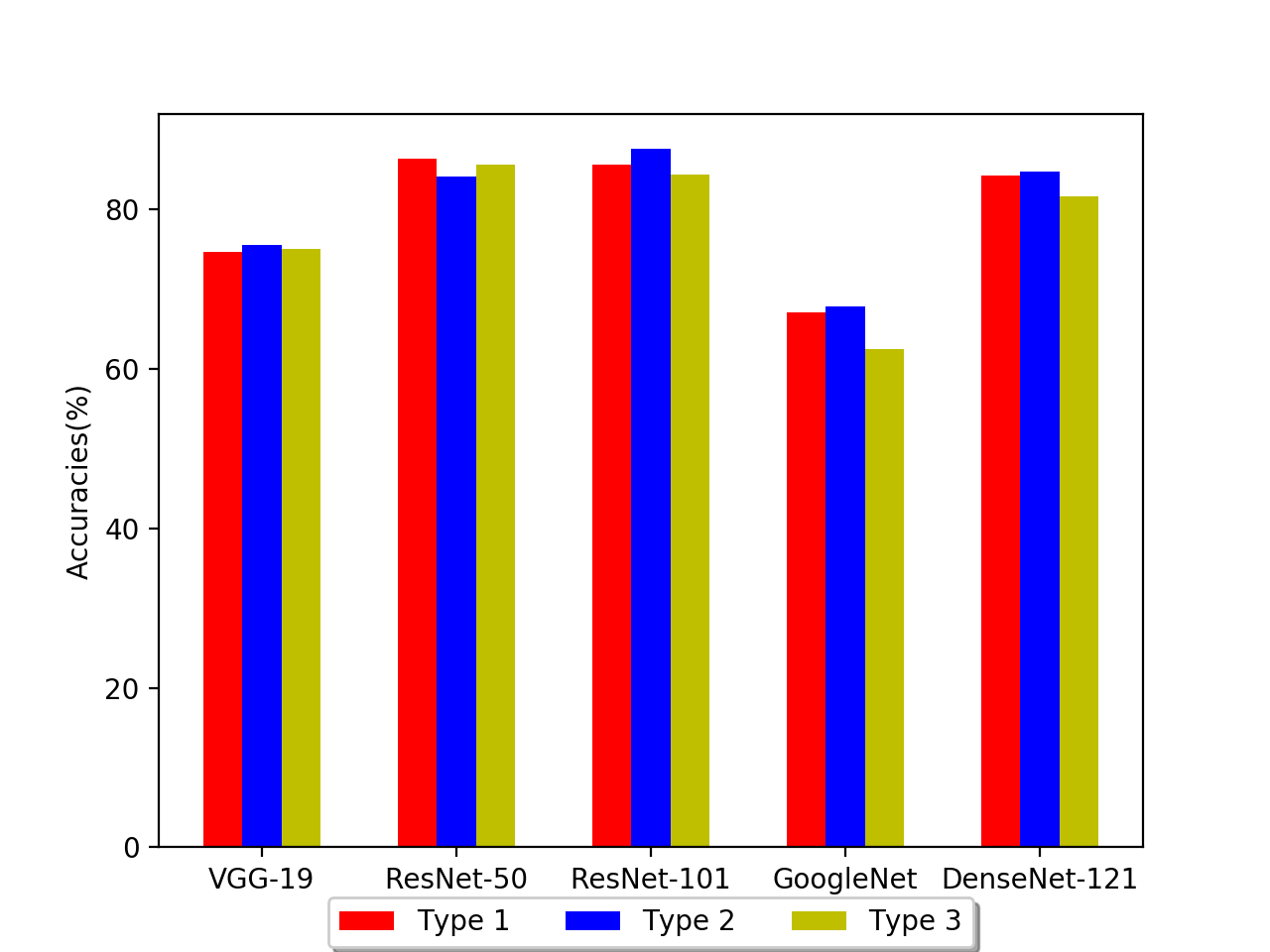}
        \caption{Effect of different artifact types on model's performance.}
        \label{fig:accvsarti}
\end{figure}

In this experiment, we investigate the impact of varying the artifact type and examine the model's performance. We notice that on an average all the models tend to result in little higher accuracy for artifact type 2. However, overall it can be seen that there is not much significant difference in model performances across different artifact types. Therefore it can be assumed that all the artifacts bring in almost same amount of noise in the image no matter how uniform or random the artifact is. Figure \ref{fig:accvsarti} shows a bar plot of accuracies for different models.

\section{Conclusion}
The experiments carried out have provided in depth  insights on the occ of objects in the data sets and helped a great deal in understanding how the models\' performances differ with variants of occs. Following are the findings, and facts we were able to establish based on our analysis:
\begin{enumerate}
    \item The type of occ or artifact added to object doesn't seem to have a significant effect on accuracy of the model. CNN based models tend to behave the same way with either random, constant or images based occ's.
    \item Based on our experiment with of the current dataset varied depth of the same network architecture(ResNet-50/101/151) we could conclude that  the depth of the networks have no effect on the predictions with occ's in anyway. 
    \item  The effect of Non pretrained model (NPT) was unable to be identified as the model was not able to generalize well due to the smaller nature of  dataset.
    \item Training the models with occluded dataset helps the these perform better when tested with datasets containing occ's. As the occ size increases the accuracy gap between with and without  occ trained models becomes wider.
    \item Though it can be understood that some models performed better than other for both the type of datasets un-occluded and occluded. However, there is not much evidence to show that if some model was performing better with regards to occ.
    \item Other interesting finding was that models trained on lower occ datasets perform better  on  higher occ datasets as compared to the model trained on unoccluded dataset. This indicates that the model is better able to generalize when it has already seen some occ in the dataset even though smaller.

    \item On the contrary, CNN models trained on higher occ datasets perform similar to models trained on un-occluded dataset when tested on  either lower occluded datasets. This also holds true when the testing is done for the un-occluded dataset. Therefore, having occ's in the training set seems to have no adverse effect on the model (which intuitively we were expecting) even though there are no occ's seen in  the original image. 
     
\end{enumerate} 
One of the future works will be to establish the findings on a real world occluded dataset to prove out theory from the artificially created dataset. Also, as we have identified earlier all CNN based models tend to behave the same when subjected to occ, the possible extension of the work can be with regards to building a model which is robust to occ's using the properties which were identified and established earlier about occ's. 
      
{\small
\bibliographystyle{ieee}
\bibliography{egbib}
}

\end{document}